\documentclass[10pt,twocolumn,letterpaper]{article}
\usepackage{sparsehps}
\usepackage{microtype}
\usepackage{multirow}
\usepackage{array}
\usepackage{tikz}
\usetikzlibrary{arrows.meta,calc,positioning,fit}

\newcommand{\family}{SparseHPS\xspace}    
\newcommand{\method}{\mbox{SparseHPS-T}\xspace}  
\newcommand{\base}{\mbox{SparseHPS-S}\xspace}    
\newcommand{\jl}{J-Err(L)\xspace}
\newcommand{\vl}{V-Err(L)\xspace}
\newcommand{\jg}{J-Err(G)\xspace}
\newcommand{\vg}{V-Err(G)\xspace}
\newcommand{\Ang}{Ang\xspace}
\newcommand{\ve}[1]{\mathbf{#1}}
\newcommand{\R}{\mathbb{R}}
\newcommand{\best}[1]{\textbf{#1}}

\newcommand{\dt}[1]{\textcolor{black!70}{\footnotesize #1}}

\definecolor{CBblue}{HTML}{0072B2}
\definecolor{CBorange}{HTML}{D55E00}
\definecolor{CBgreen}{HTML}{009E73}
\definecolor{CBconv}{HTML}{E8EEF4}
\definecolor{CBattn}{HTML}{CFE2F3}
\definecolor{CBdec}{HTML}{FBE3D0}
\definecolor{CBdecdark}{HTML}{B85C00}
\definecolor{CBshape}{HTML}{009E73}
\definecolor{CBtransl}{HTML}{D55E00}
\definecolor{CBjoint}{HTML}{0072B2}

\newcommand{\dn}{\,$\downarrow$}                 
\newcommand{\fxp}[2]{#1\,$\times$\,#2}           

\definecolor{linkblue}{rgb}{0.21,0.49,0.74}
\usepackage[breaklinks,colorlinks,allcolors=linkblue]{hyperref}

\newcommand{\papertitle}{SparseHPS: A Unified Model for Human Pose and Shape Estimation\texorpdfstring{\\}{ }across Depth Cameras and LiDARs}
\newcommand{\authorname}{Ilya Zaychuk}
\title{\papertitle}
\author{\authorname\\
  \href{https://www.linkedin.com/in/ilya-zaychuk}{\nolinkurl{linkedin.com/in/ilya-zaychuk}}}
\hypersetup{pdftitle={\papertitle},pdfauthor={\authorname}}

\begin{document}
\maketitle
\begin{abstract}
Recovering 3D human pose and shape from point clouds is central to motion capture, robotics and self-driving vehicles, because depth cameras and LiDARs sense a person directly in metric 3D and work under any lighting.
However, current approaches are each tailored to one sensor type: point density differs by more than ten times between devices, public datasets are labeled with incompatible body models, and LiDAR approaches need long input sequences to offset sparse scans.
In this work, we present \family, a novel sensor-agnostic network that recovers SMPL-X pose, shape and global translation from a human point cloud with one set of weights, for single frames and streams alike.
Specifically, \family encodes the cloud with a LitePT backbone over sparse voxels, represents every joint by a learnable query decoded with one common head, localizes the body by voting over voxels, and draws temporal context from a causal memory of tokens from the seven preceding frames.
A single model is trained on ten datasets that mix rendered clouds with real depth-camera and LiDAR captures, all converted to one body model.
Experiments demonstrate state-of-the-art results on the SLOPER4D and FreeMotion LiDAR benchmarks, where \family lowers the root-relative joint error by 22\% and 11\%, respectively, compared with the strongest prior method, which needs 32 frames.
On real Azure Kinect captures of held-out HuMMan subjects, \family attains a joint error of 34.7\,mm.
Moreover, the current frame alone already yields the bulk of this accuracy.\\
Project page: \href{https://ghostlate.github.io/SparseHPS}{\nolinkurl{ghostlate.github.io/SparseHPS}}\par\vspace*{-12pt}
\end{abstract}

\section{Introduction}
\label{sec:intro}

\begin{figure}[t]
  \centering
  \setlength{\tabcolsep}{0pt}
  \newcommand{\tq}[3]{\shortstack{{\tiny\makebox[0.775cm]{input}\makebox[0.775cm]{ours}\makebox[0.775cm]{GT}}\\[1pt]#1#2#3}}
  \begin{tabular}{@{}c@{\hspace{3pt}}c@{\hspace{3pt}}c@{}}
    \tq{\includegraphics[height=1.55cm]{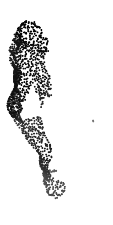}}{\includegraphics[height=1.55cm]{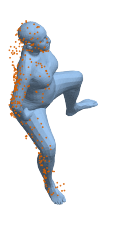}}{\includegraphics[height=1.55cm]{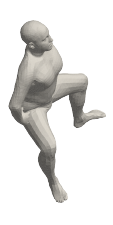}}
    & \tq{\includegraphics[height=1.55cm]{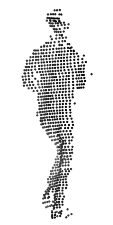}}{\includegraphics[height=1.55cm]{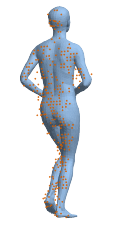}}{\includegraphics[height=1.55cm]{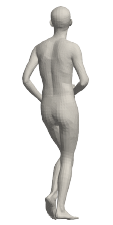}}
    & \tq{\includegraphics[height=1.55cm]{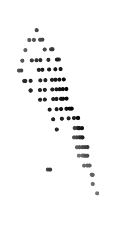}}{\includegraphics[height=1.55cm]{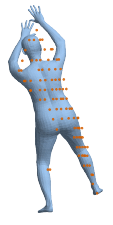}}{\includegraphics[height=1.55cm]{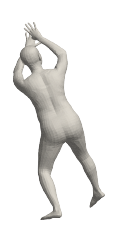}} \\[-2pt]
    {\footnotesize depth camera, 1.7k points} & {\footnotesize LiDAR, 0.8k points} & {\footnotesize LiDAR, 90 points} \\[2pt]
  \end{tabular}
  \includegraphics[width=\linewidth]{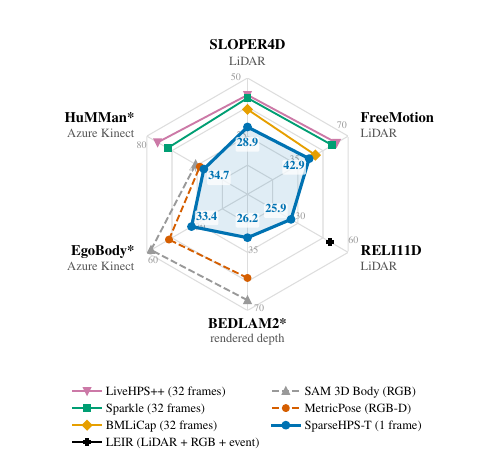}
  \caption{\textbf{A single model for depth-camera and LiDAR clouds.}
    \emph{Top:} test samples from HuMMan, SLOPER4D and FreeMotion, each with an error typical of its set: the input after 2\,cm voxelization, the single-frame output of \family (blue, with the input in orange) and the ground truth (GT, grey).
    \emph{Bottom:} root-relative joint errors (mm) on six benchmarks (\cref{tab:lidar,tab:humman,tab:reli,tab:b2,tab:egobody}); each line is a method, drawn on the benchmarks it reports, and each axis runs from 0 at the centre to the value at its end, so that lower errors lie closer to the centre.
    \method from the current frame only; the inputs of the other methods are given in brackets.
    $^*$The published results use other splits than ours.}
  \label{fig:teaser}
\end{figure}

Estimating the 3D pose and shape of the human body is a long-standing task in computer vision that underpins motion capture, robotics and autonomous driving.
Unlike RGB images, depth cameras and LiDARs deliver metric 3D measurements of the body: they remove the ambiguity of depth and scale, are unaffected by illumination, and place the person in world coordinates.

However, existing point-based methods are designed around one kind of sensor.
Depth-camera approaches~\cite{jiang2019skeleton,wang2020sequential,liu2021votehmr,cai2023pointhps} expect dense inputs of several thousand sampled points, while LiDAR approaches~\cite{li2022lidarcap,ren2024livehps,ren2024livehpspp,ren2026sparkle,an2026bmlicap} operate on sparse scans with only hundreds of points and recover the missing geometry by aggregating long point sequences, up to 32 frames in the latest methods.
Each sensor has its own models and benchmarks, and the data are not interchangeable: after 2\,cm voxelization, the median person has 2.3k points in the Azure Kinect views of HuMMan~\cite{cai2022humman} but about 100 for the most distant LiDAR in FreeMotion~\cite{ren2024livehps}; in addition, labels come in different body models.
Image-based mesh recovery, by comparison, has gained substantially from training one model across many datasets~\cite{goel2023hmr2,cai2023smplerx}.
We are not aware of any model that serves both sensor families with one set of weights.

\begin{figure*}[t]
  \centering
  \input{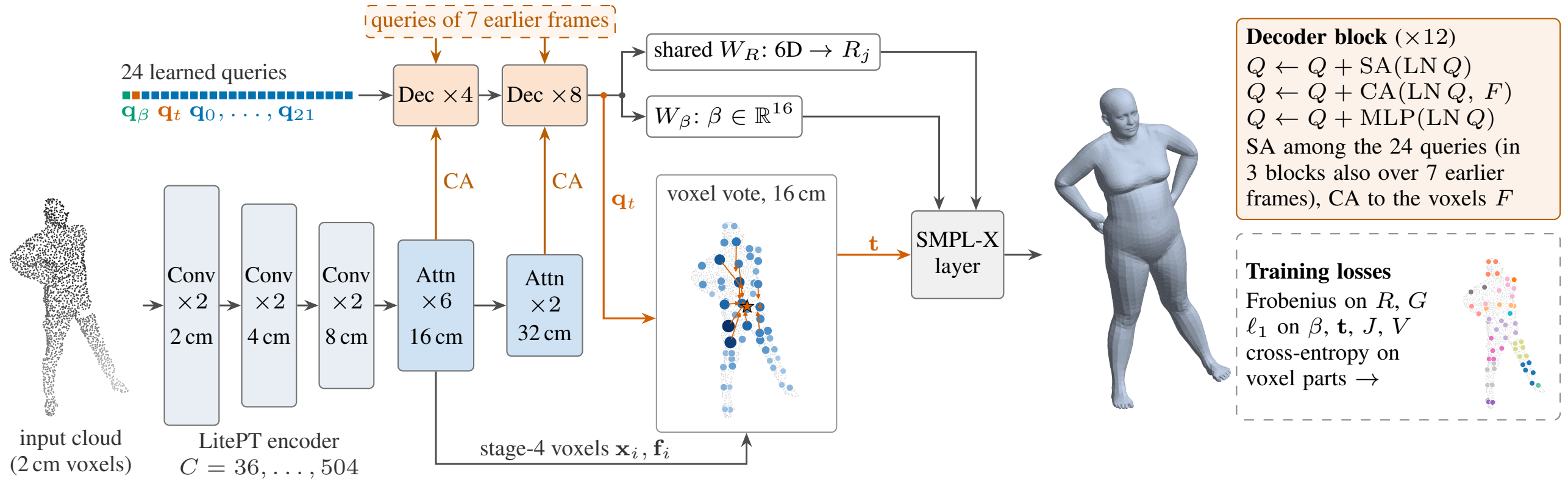}
  \caption{\textbf{Architecture of \family.}
    A LitePT encoder processes the voxelized cloud with convolutions in stages 1--3 and attention in stages 4--5.
    In the two coarsest stages, 24 learned queries (shape, translation and one per body joint) alternate self-attention with cross-attention (CA) to the voxels of their own cloud; in three of the twelve decoder blocks, every query additionally attends to its own values in up to seven previous frames.
    A shared linear layer maps every joint query to a 6D rotation.
    The translation query weights the stage-4 voxels, each of which votes for the pelvis with an offset.
    Panels: a forward pass on a held-out HuMMan subject; voxel colour is the total vote weight $\sum_h\pi_h a_{ih}$, arrows are the votes of the 12 highest-weight voxels, the star is the predicted pelvis.
    The part segmentation of the stage-4 voxels (bottom right, predicted) is only a training loss.}
  \label{fig:arch}
\end{figure*}

To bridge this gap, we present \family (\cref{fig:arch}), a novel network that estimates human pose and shape from point clouds, frame by frame or over a stream.
Its encoder is LitePT~\cite{yue2026litept}, a point transformer over sparse voxels that uses convolutions at the fine levels and attention at the coarse ones.
Specifically, 24 learnable query tokens, one per body joint plus one for shape and one for translation, cross-attend to voxels from the two coarsest stages of the encoder, and all joint tokens share one linear head.
Instead of predicting the global position directly from a token, each voxel votes for the pelvis location, weighted by the translation token.
In addition, a causal memory lets each token attend to its own states in the seven previous frames; for a frame with no history, the computation reduces exactly to the single-frame model.
We refer to the single-frame variant as \base and to the variant with memory as \method.

To train one model for all sensors, we bring ten sources into a common label space, the neutral SMPL-X model with 16 shape coefficients: three LiDAR datasets (SLOPER4D, FreeMotion, RELI11D), four depth-camera datasets (HuMMan, BEHAVE, EgoBody, FORM-HOI), and renderings of synthetic and scanned humans (AMASS, BEDLAM2, CAPE).
Per-source quotas keep large datasets with many views from dominating training, and a rotation augmentation that respects gravity retains the vertical direction on which sparse scans depend.

Extensive experiments on the full mixture support every design decision.
Over seven consecutive single-frame models, each changing a single component, the vertex error averaged over nine test sets falls from 43.3\,mm to 36.0\,mm (\cref{tab:ladder}); a data ladder further shows that every sensor family needs data of its own, whereas adding the other families comes at almost no cost (\cref{tab:data_ladder}).
The memory leaves the mean single-frame accuracy intact and cuts acceleration error by 5\% (\cref{tab:memory}).
On the LiDAR benchmarks, \family sets a new state of the art: with a history of seven frames, its root-relative joint error is 28.6\,mm (SLOPER4D) and 41.8\,mm (FreeMotion), whereas the strongest published method obtains 36.5\,mm and 47.2\,mm from 32 frames (\cref{fig:teaser}).
Notably, a single frame already reaches 28.9 and 42.9\,mm, since the encoder sees every point of the scan.

In summary, our contributions are threefold:
\begin{itemize}
  \item We introduce \textbf{\family}, a novel one-stage model that recovers SMPL-X pose, shape and global position from single-frame or sequential point clouds, using per-joint query tokens with a shared decoding head, a voting scheme over voxels and a causal memory of tokens.
  \item We develop \textbf{a training recipe that spans range sensors}: ten sources mapped to one label space, per-source quotas and a gravity-preserving augmentation, yielding shared weights across depth cameras and LiDARs.
  \item We provide \textbf{controlled ablations on the complete training mixture} and reach state-of-the-art accuracy on LiDAR benchmarks with one quarter of the frames that sequence-based methods take.
\end{itemize}

\section{Related Work}
\label{sec:related}

\paragraph{Human mesh recovery from images.}
Parametric body models such as SMPL~\cite{loper2015smpl} and SMPL-X~\cite{pavlakos2019smplx} are regressed from an image with a pooled feature and an iterative head~\cite{kanazawa2018hmr,kolotouros2019spin,zhang2021pymaf}, or with query tokens that attend to image features, per joint and vertex~\cite{lin2021metro,cho2022fastmetro} or per person~\cite{goel2023hmr2,baradel2024multihmr,dwivedi2024tokenhmr}, following detection queries~\cite{carion2020detr}.
Recent progress is largely driven by data: large synthetic datasets~\cite{black2023bedlam,tesch2025bedlam2} and joint training on many datasets~\cite{cai2023smplerx,ranftl2022midas}, also with heterogeneous annotations~\cite{sarandi2024nlf}.
Video-based methods further exploit temporal context~\cite{kanazawa2019hmmr,kocabas2020vibe}, and gravity-aligned coordinates facilitate global motion recovery~\cite{shen2024gvhmr}.

\paragraph{Depth cameras and scans.}
Early depth-based approaches label pixels with body parts~\cite{shotton2011realtime}, an idea that we adopt as an auxiliary loss on voxels.
Subsequent methods regress SMPL parameters from PointNet++ features of partial point clouds~\cite{jiang2019skeleton,wang2020sequential}; VoteHMR~\cite{liu2021votehmr} lets points vote for joints, inspired by Hough voting for object detection~\cite{qi2019votenet}, and PointHPS~\cite{cai2023pointhps} refines point features in a cascaded architecture.
In contrast, we employ voting only to localize the body and estimate the pose with query tokens.
Another line of work fits a body model to complete scans, through predicted correspondences and optimization~\cite{bhatnagar2020ipnet,bhatnagar2020loopreg,wang2021ptf,li2025etch,li2026etchx} or equivariant regression~\cite{feng2023arteq}, whereas we target single-view sensor data and a single forward pass.

\paragraph{LiDAR-based motion capture.}
LiDARCap~\cite{li2022lidarcap} pioneered long-range motion capture from LiDAR point sequences, and LiDAR-HMR~\cite{fan2023lidarhmr} reconstructs the mesh from a single scan.
LiveHPS~\cite{ren2024livehps} and LiveHPS++~\cite{ren2024livehpspp} estimate pose, shape and translation from windows of 32 frames, each subsampled to 256 points, in free environments with noise and occlusion.
Sparkle~\cite{ren2026sparkle} proposes a representation that is evaluated on LiDAR and depth-camera benchmarks with a separate model for each, and BMLiCap~\cite{an2026bmlicap} models the degradation of sparse scans.
In autonomous driving, 3D keypoints of pedestrians are estimated from LiDAR alone~\cite{ye2024lpformer} or together with images~\cite{zanfir2022hum3dil}.
In contrast to these sensor-specific and mostly sequence-based methods, we show that a single model trained across sensor families is more accurate on their benchmarks already from a single frame.

\paragraph{Point cloud backbones.}
PointNet++~\cite{qi2017pointnet,qi2017pointnetpp}, the encoder of most methods above, operates on a few hundred to a few thousand sampled points.
3D scene understanding has moved to sparse voxel convolutions~\cite{graham2018submanifold,choy2019minkowski} and point transformers~\cite{zhao2021pt,wu2022ptv2,wu2024ptv3}, which scale to all points of a scan; we adopt the encoder of LitePT~\cite{yue2026litept}, which applies convolutions at fine resolutions and attention at coarse ones.

\section{Method}
\label{sec:method}

\paragraph{Problem setting.}
The input is the point cloud $P=\{\ve{p}_k\}_{k=1}^N\subset\R^3$ of one person, in metres, and optionally the clouds of the same person in previous frames.
We assume that the vertical axis of the input frame is known up to a tilt of about $15^\circ$, \eg from an IMU, the sensor calibration or the ground plane; the yaw of the person is unknown.
We translate $P$ to the centre $\ve{c}$ of its bounding box and voxelize it at 2\,cm, keeping one point per voxel; the point coordinates are the only input features.
The output is a body of the SMPL-X model~\cite{pavlakos2019smplx} in the neutral locked-head variant of BEDLAM2~\cite{tesch2025bedlam2}: 16 shape coefficients $\beta$, the rotations $R_0,\dots,R_{21}\in SO(3)$ of the root and of the 21 body joints, each relative to its parent, and the position $\ve{t}$ of the pelvis joint in the centred frame, so that the pelvis in the sensor frame is $\ve{t}+\ve{c}$.

\subsection{Sparse-voxel encoder}
\label{sec:method:backbone}
We adopt the encoder of LitePT~\cite{yue2026litept}, a five-stage point transformer on sparse voxels, and discard its U-Net decoder, since a body model has no per-point output.
Each stage halves the resolution by grid max-pooling, which yields cells of 2, 4, 8, 16 and 32\,cm with 36, 72, 144, 252 and 504 channels.
Stages 1--3 consist of two submanifold sparse convolution blocks~\cite{graham2018submanifold} each, whereas stages 4 and 5 consist of 6 and 2 pre-norm transformer blocks, which attend within patches of up to 1024 voxels serialized along space-filling curves~\cite{wu2024ptv3} and use rotary embeddings~\cite{su2024roformer} of the grid positions.

This design accommodates the density gap between sensors: at 2\,cm, the median cloud contains 2.0--2.7k voxels for the depth-camera datasets, 0.8k for SLOPER4D and 0.26k for FreeMotion, whereas at stage 4 every sensor is reduced to 45--75 voxels of 16\,cm, few enough for a single attention patch to cover the whole person and fine enough to separate the limbs.

\subsection{Per-joint query decoder}
\label{sec:method:tokens}
Instead of regressing all parameters from one pooled feature, we introduce $T=24$ learned query tokens $Q=[\ve{q}_\beta,\ve{q}_t,\ve{q}_0,\dots,\ve{q}_{21}]\in\R^{T\times 252}$, shared across inputs: one for the shape, one for the translation and one for each body joint.
The tokens pass through $D_4=4$ decoder blocks after the encoder blocks of stage 4, are projected to 504 channels, and pass through $D_5=8$ further blocks after the encoder blocks of stage 5.
Given the voxel features $F$ of a cloud at the respective stage, a decoder block updates the tokens of this cloud as
\begin{align}
  Q &\leftarrow Q + \mathrm{SA}(\mathrm{LN}(Q)), \nonumber\\
  Q &\leftarrow Q + \mathrm{CA}(\mathrm{LN}(Q), F), \label{eq:dec}\\
  Q &\leftarrow Q + \mathrm{MLP}(\mathrm{LN}(Q)), \nonumber
\end{align}
where SA denotes self-attention among the 24 tokens, which exchanges information between joints, and CA denotes cross-attention from the tokens to all voxels of the cloud.

\paragraph{Prediction heads.}
The shape is predicted from the shape token as $\beta=\bar\beta+W_\beta\ve{q}_\beta$, where $\bar\beta$ is the mean training shape.
All joint tokens are decoded by the \emph{same} linear layer into a 6D rotation~\cite{zhou2019continuity}, $R_j=\mathrm{GS}(\bar{\ve{r}}_j+W_R\ve{q}_j)$, where GS denotes Gram--Schmidt orthonormalization and $\bar{\ve{r}}_j$ is the mean rotation of joint $j$.
With a shared $W_R$, the head has no joint-specific parameters, and the identity of a joint is encoded by its query alone.

\subsection{Translation by voxel voting}
\label{sec:method:vote}
The offset of the pelvis from the bounding-box centre depends on the pose and on the visible part of the body, whereas its position relative to a voxel is a local quantity that is largely determined by the surrounding surface.
Instead of regressing $\ve{t}$ from a token, we therefore let the stage-4 voxels vote for the pelvis.
For voxel $i$ with feature $\ve{f}_i$ and centroid $\ve{x}_i$, and for each of $H=4$ heads $h$, we compute
\begin{align}
  \ell_{ih} &= \tfrac{1}{\sqrt d}\,\big\langle W^q_h\ve{q}_t,\; W^k_h\,\mathrm{LN}(\ve{f}_i)\big\rangle + g_h(\ve{x}_i-\bar{\ve{x}}), \label{eq:vote_logit}\\
  \ve{v}_{ih} &= \ve{x}_i + 0.5\,\tanh\!\big(W^o_h\,\mathrm{LN}(\ve{f}_i)\big), \label{eq:vote_offset}\\
  \ve{t} &= \textstyle\sum_h \pi_h \sum_i a_{ih}\,\ve{v}_{ih} + W_r\ve{q}_t, \label{eq:vote}
\end{align}
where $a_{\cdot h}=\mathrm{softmax}_i(\ell_{ih})$ is normalized over the voxels of one cloud, $\pi=\mathrm{softmax}(W^\pi\ve{q}_t)$, $d=32$, and $g_h$ is a small MLP of the voxel position relative to the mean voxel position $\bar{\ve{x}}$ of the cloud.
Each voxel thus proposes a pelvis position $\ve{v}_{ih}$ within 0.5\,m of itself, the translation token weights the proposals, and the residual $W_r\ve{q}_t$ accounts for what voting cannot express.
We zero-initialize $W^q$, $W^o$, $W^\pi$, $W_r$ and the last layer of $g_h$, so that training starts from the centroid of the voxels; the votes are supervised only through the translation loss.

\subsection{Causal token memory}
\label{sec:method:memory}
For streaming input, every fourth decoder block (three of the twelve) is equipped with a causal memory, following the space--time factorization of MEM~\cite{torne2026mem}.
Let $\ve{a}_{n,j}$, $\ve{k}_{n,j}$ and $\ve{v}_{n,j}$ denote the query, key and value projections of token $j$ in frame $n$ in the self-attention layer of such a block, and let $\mathcal{H}_n$ be the set of up to $K=7$ previous frames.
The value of each token first attends over the \emph{same} token across time, and the self-attention among the tokens of frame $n$ then operates on the aggregated values:
\begin{align}
  \hat{\ve{v}}_{n,j} &= \!\!\textstyle\sum\limits_{m\in\mathcal{H}_n\cup\{n\}}\!\!\! \mathrm{softmax}_{m}\big(\langle\ve{a}_{n,j},\ve{k}_{m,j}\rangle + b\,[m{\neq}n]\big)\,\ve{v}_{m,j}, \nonumber\\
  \ve{o}_{n,i} &= \textstyle\sum_{j} \mathrm{softmax}_{j}\big(\langle\ve{a}_{n,i},\ve{k}_{n,j}\rangle\big)\,\hat{\ve{v}}_{n,j}. \label{eq:memory}
\end{align}
Keys and values of a previous frame are obtained with the existing projections from its token plus an MLP embedding of the relation between the two frames, namely their time difference and the rigid transformation between their coordinate frames, which is required because every cloud is centred individually.
This MLP and one bias $b$ per attention head are the only additional parameters (0.6M in total).
A frame without a history undergoes exactly the computation of a network without memory; hence, a single set of weights handles single clouds and streams, and the memory can be added to a trained single-frame model (\cref{sec:training}).
As the memory only reads tokens of previous frames, they can be cached, so that a frame of a stream costs one forward pass.

\subsection{Training objectives}
\label{sec:method:loss}
The SMPL-X layer converts the prediction into joints $J$ and vertices $V$.
The network is trained with
\begin{equation}
  \mathcal{L}=\mathcal{L}_R+\mathcal{L}_\beta+\mathcal{L}_t+\mathcal{L}_J+\mathcal{L}_V+\lambda_G\mathcal{L}_G+\lambda_S\mathcal{L}_S ,
  \label{eq:loss}
\end{equation}
where $\lambda_G=0.1$ and $\lambda_S=3$.
$\mathcal{L}_\beta$ and $\mathcal{L}_t$ are $\ell_1$ losses, $\mathcal{L}_J$ is the mean $\ell_1$ error over 42 joints (the 22 body joints, jaw, nose, eyes, 6 foot points and 10 fingertips), and $\mathcal{L}_V$ is the mean $\ell_1$ error over all 10{,}475 vertices.
For rotations, we penalize the Frobenius norm of the residual of each joint with respect to the ground truth $\hat R_j$,
\begin{equation}
  \begin{aligned}
    \mathcal{L}_R&=\kappa\textstyle\sum_{j=0}^{21}\lVert \hat R_j-R_j\rVert_F, \\
    \mathcal{L}_G&=\kappa\textstyle\sum_{j=0}^{21}\lVert \hat G_j-G_j\rVert_F,
  \end{aligned}
  \label{eq:rot}
\end{equation}
where $G_j=G_{\mathrm{pa}(j)}R_j$ are the global rotations obtained by forward kinematics.
For an error of angle $\theta$, $\lVert \hat R-R\rVert_F=2\sqrt2\sin(\theta/2)$; the loss thus depends only on the error angle, whereas an element-wise $\ell_1$ loss on the matrix also depends on the rotation axis, and $\kappa=2.4$ preserves the scale of that $\ell_1$ loss.
$\mathcal{L}_G$ additionally penalizes errors that accumulate along the kinematic chain.

\paragraph{Voxel part segmentation.}
As auxiliary supervision, a linear layer on the stage-4 voxel features predicts one of 23 classes: 22 body parts, defined by the dominant skinning weight of each vertex (fingers are assigned to the wrists and the face to the head), or background.
The label of a voxel is the part of the nearest of 2{,}000 ground-truth vertices selected by farthest point sampling, or background if this vertex is farther than 5\,cm, and $\mathcal{L}_S$ is the cross-entropy over the voxels.
This task encourages every voxel to encode the body part it lies on, which is the information that the joint tokens query; the head is discarded at test time.

\section{Unified Training Across Sensors}
\label{sec:training}

\begin{table}[t]
  \centering
  \small
  \setlength{\tabcolsep}{3pt}
  \resizebox{\linewidth}{!}{%
  \begin{tabular}{@{}llrrr@{}}
    \toprule
    Source & Sensor & Clouds & People & Share \\
    \midrule
    BEDLAM2~\cite{tesch2025bedlam2} & rendered depth, clothed & 575k & 1{,}054 & 40.3\% \\
    AMASS~\cite{mahmood2019amass} & virtual Kinect, unclothed & 387k & 425\textsuperscript{a} & 33.1\% \\
    CAPE~\cite{ma2020cape} & virtual Kinect, clothed scans & 32k & 12 & 2.3\% \\
    \midrule
    HuMMan~\cite{cai2022humman} & 10$\times$ Azure Kinect & 421k & 295 & 3.0\% \\
    EgoBody~\cite{zhang2022egobody} & 5$\times$ Azure Kinect & 168k & 28 & 3.3\% \\
    BEHAVE~\cite{bhatnagar2022behave} & 4$\times$ Azure Kinect & 45k & 7 & 1.3\% \\
    FORM-HOI~\cite{nvidia2026formhoi} & 4$\times$ stereo depth & 1{,}640k & 13 & 9.0\% \\
    \midrule
    FreeMotion~\cite{ren2024livehps} & 3$\times$ LiDAR, static & 98k & -- & 3.6\% \\
    SLOPER4D~\cite{dai2023sloper4d} & LiDAR, moving & 23k & 5 & 2.2\% \\
    RELI11D~\cite{yan2024reli11d} & LiDAR, static & 22k & 6 & 2.0\% \\
    \bottomrule
  \end{tabular}}
  \caption{\textbf{Training sources.}
    A cloud is one person seen by one sensor in one frame.
    \emph{Share}: fraction of an epoch of the single-frame stage (1.77M clouds).
    The ablations (\cref{sec:exp:ablation,sec:exp:data}) use the sources other than CAPE.
    \textsuperscript{a}Subject folders; 75\% of the AMASS clouds use a randomly drawn body shape.}
  \label{tab:data}
\end{table}

The architecture of \cref{sec:method} contains no sensor-specific component, so generalization across sensors has to come from the data.
We unify ten sources that differ in sensor, clothing, subjects and annotation (\cref{tab:data}), balance them, and train in two stages.

\begin{table*}[t]
  \centering
  \small
  \setlength{\tabcolsep}{5pt}
  \resizebox{\textwidth}{!}{%
  \begin{tabular}{@{}llcccccc@{}}
    \toprule
    & & \multicolumn{3}{c}{SLOPER4D} & \multicolumn{3}{c}{FreeMotion} \\
    \cmidrule(lr){3-5}\cmidrule(l){6-8}
    Method & Input (frames\,$\times$\,points) & \jl/\vl\dn & \jg/\vg\dn & \Ang\dn & \jl/\vl\dn & \jg/\vg\dn & \Ang\dn \\
    \midrule
    LiDARCap~\cite{li2022lidarcap}\textsuperscript{a} & sequence & 71.6/84.2 & 138.7/147.8 & 13.72 & 86.3/104.2 & 180.4/188.6 & 15.51 \\
    LiveHPS~\cite{ren2024livehps} & \fxp{32}{256} & 48.3/59.0 & 77.7/85.8 & 12.77 & 69.4/83.9 & 119.3/128.6 & 15.79 \\
    LiveHPS++~\cite{ren2024livehpspp} & \fxp{32}{256} & 42.7/50.6 & 77.0/81.7 & 11.92 & 61.9/75.3 & 112.1/120.4 & 15.40 \\
    Sparkle~\cite{ren2026sparkle} & \fxp{32}{256} & 41.4/50.5 & 70.9/77.1 & 10.48 & 59.0/73.2 & 105.1/113.9 & 9.66 \\
    BMLiCap~\cite{an2026bmlicap} & sequence & 39.7/47.8 & -- & -- & 49.6/60.3 & -- & -- \\
    BMLiCap\textsuperscript{\dag}~\cite{an2026bmlicap} & 32 frames & 36.5/44.2 & -- & -- & 47.2/59.0 & -- & -- \\
    \midrule
    LiveHPS~\cite{ren2024livehps}, released model, our evaluation\textsuperscript{b} & \fxp{32}{256} & 47.4/57.9 & 115.6/123.3 & 10.88 & 52.8/63.7 & 96.3/102.6 & 9.22 \\
    \midrule
    \base, synthetic sources only & \fxp{1}{all} & 50.7/62.8 & 102.6/111.1 & 12.23 & 86.9/107.5 & 122.8/134.1 & 14.39 \\
    \base, no LiDAR sources & \fxp{1}{all} & 48.9/60.6 & 97.9/105.7 & 12.66 & 82.6/102.6 & 120.8/131.7 & 13.51 \\
    Single-token baseline & \fxp{1}{all} & 31.3/38.3 & 75.8/81.1 & 7.95 & 49.8/60.8 & 80.9/87.8 & 7.33 \\
    \base & \fxp{1}{all} & 29.0/35.8 & 69.6/75.3 & 7.80 & 42.4/52.8 & 66.9/73.9 & 6.94 \\
    \midrule
    \method, 256 points\textsuperscript{d} & \fxp{8}{256} & 36.8/45.5 & 79.3/86.3 & 9.91 & 48.2/60.0 & 76.9/84.4 & 8.28 \\
    \method, no history & \fxp{1}{all} & 28.9/35.8 & 69.8/75.5 & 7.83 & 42.9/53.3 & 68.2/75.2 & 6.96 \\
    \method & \fxp{8}{all} & \best{28.6}/\best{35.3} & \best{69.4}/\best{75.0}\textsuperscript{c} & \best{7.76} & \best{41.8}/\best{51.8} & \best{66.8}/\best{73.5} & \best{6.82} \\
    \bottomrule
  \end{tabular}}
  \caption{\textbf{Comparison on LiDAR benchmarks} under the LiveHPS protocol (SMPL, mm; \Ang in degrees).
    First block: published results from the paper of each method (BMLiCap reports only local errors; \dag: its 32-frame variant).
    Third block: single-frame models of our ablations (\cref{tab:data_ladder,tab:ladder}); the first two have never seen a LiDAR scan.
    \method processes the current frame and 7 previous frames 0.2\,s apart, causally.
    \textsuperscript{a}As retrained in~\cite{ren2024livehpspp}.
    \textsuperscript{b}Released code and weights, official script, on our test files (\cref{sec:exp:setup}).
    \textsuperscript{c}26 frames (0.5\%) contain fewer than 20 points, stray returns metres away from the person; without them \jg/\vg is 30.8/36.0.
    \textsuperscript{d}The same weights on clouds reduced to 256 points by farthest point sampling.}
  \label{tab:lidar}
\end{table*}

\paragraph{Unified label space.}
The sources are annotated with SMPL~\cite{loper2015smpl} (HuMMan, FreeMotion, SLOPER4D, RELI11D, CAPE), SMPL-H (BEHAVE), SMPL-X (EgoBody) or MHR~\cite{ferguson2025mhr} (FORM-HOI), often with gendered models and 10 shape coefficients, whereas BEDLAM2 and AMASS already use the neutral locked-head SMPL-X model with 16 coefficients.
We refit all other sources to this model by optimizing pose, shape and translation against the source mesh converted to the SMPL-X topology; the mean vertex error of the refit is 1.5--6.0\,mm per source.

\paragraph{Sources and splits.}
BEDLAM2 provides depth renderings of clothed people captured by moving cameras; as it replays the same motions on many bodies, we exclude from training every track whose poses are within $2^\circ$ of a BEDLAM2 validation frame or of the AMASS evaluation sets.
AMASS contributes pose diversity: we ray-cast its body meshes from a virtual Azure Kinect and, for 75\% of the clouds, replace the body shape by a random one, which decorrelates pose and shape.
CAPE contributes real clothing in motion; its registered 4D scans are ray-cast in the same way.
We split HuMMan by subject (295 for training, 42 for testing), since a per-cloud split would place the 10 views of a labelled frame on both sides, and hold out 9 subjects of FORM-HOI.
EgoBody, BEHAVE, RELI11D and CAPE keep their standard splits, and FreeMotion and SLOPER4D follow LiveHPS~\cite{ren2024livehps}.
All clouds with at least 50 points are used for training.

\paragraph{Balanced sampling.}
The raw union is highly imbalanced: FORM-HOI accounts for 48\% of all clouds and BEHAVE for 1\%.
We assign each source other than BEDLAM2 and AMASS a fixed number of draws per epoch, 1.25--2 per labelled frame (0.4 for FORM-HOI, whose 410k frames show only 13 people), and let these two sources share the remaining draws in proportion to the square root of their size~\cite{arivazhagan2019massively}, which amounts to three quarters of every epoch.

\paragraph{Gravity-aware augmentation.}
With probability 0.85, we rotate the cloud about the vertical axis by a uniform angle and then tilt the vertical axis by up to $15^\circ$ in a random direction.
The tilt covers calibration errors and sloped ground, while the direction of gravity, which disambiguates sparse scans, is preserved; uniform rotations over $SO(3)$ discard it (\cref{sec:exp:ablation}).
Before the rotation, we apply point dropout, Gaussian jitter and outlier points and, for depth clouds with probability 0.05, subsampling to 250--2000 points, with strengths set per sensor group.

\begin{figure*}[t]
  \centering
  \setlength{\tabcolsep}{0pt}
  \newcommand{\qn}[1]{\makebox[1.064cm]{#1}}
  \newcommand{\qh}[3]{\shortstack{{\scriptsize\qn{input}\qn{ours}\qn{GT}}\\[1pt]#1\hspace{-1pt}#2\hspace{-1pt}#3}}
  \newcommand{\qm}[1]{\makebox[1.008cm]{#1}}
  \newcommand{\qc}[4]{\shortstack{{\scriptsize\qm{input}\qm{LiveHPS}\qm{ours}\qm{GT}}\\[1pt]#1\hspace{-1pt}#2\hspace{-1pt}#3\hspace{-1pt}#4}}
  \newcommand{\qv}[3]{\begin{tabular}[b]{@{}r@{\hspace{3pt}}c@{}}
      {\scriptsize input} & \raisebox{-0.35\height}{#1} \\[1pt]
      {\scriptsize ours} & \raisebox{-0.35\height}{#2} \\[1pt]
      {\scriptsize GT} & \raisebox{-0.35\height}{#3}
    \end{tabular}}
  \begin{tabular}{@{}c@{\hspace{5pt}}c@{\hspace{5pt}}c@{\hspace{5pt}}c@{}}
    \qh{\includegraphics[height=1.9cm]{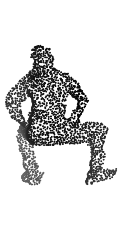}}{\includegraphics[height=1.9cm]{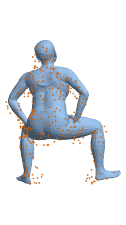}}{\includegraphics[height=1.9cm]{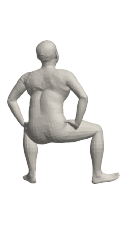}}
    & \qh{\includegraphics[height=1.9cm]{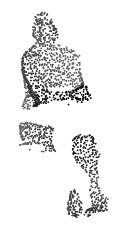}}{\includegraphics[height=1.9cm]{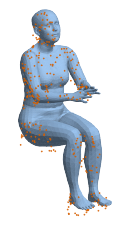}}{\includegraphics[height=1.9cm]{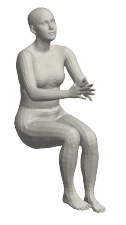}}
    & \qh{\includegraphics[height=1.9cm]{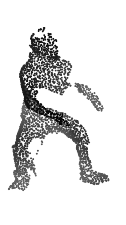}}{\includegraphics[height=1.9cm]{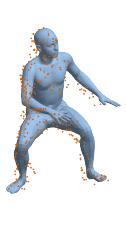}}{\includegraphics[height=1.9cm]{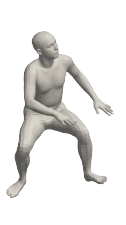}}
    & \qv{\includegraphics[width=1.63cm]{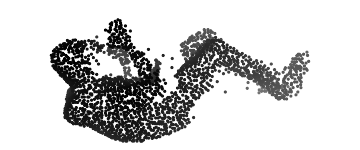}}{\includegraphics[width=1.63cm]{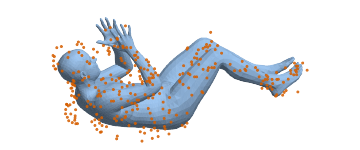}}{\includegraphics[width=1.63cm]{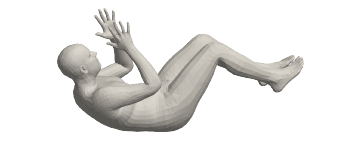}} \\[-3pt]
    {\footnotesize HuMMan, 2.6k pts, 34\,mm} & {\footnotesize EgoBody, 1.5k pts, 35\,mm} & {\footnotesize BEHAVE, 2.9k pts, 28\,mm} & {\footnotesize FORM-HOI, 2.4k pts, 29\,mm} \\[2pt]
  \end{tabular}\\
  \begin{tabular}{@{}c@{\hspace{4pt}}c@{\hspace{4pt}}c@{\hspace{4pt}}c@{}}
    \qc{\includegraphics[height=1.8cm]{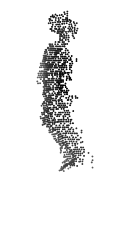}}{\includegraphics[height=1.8cm]{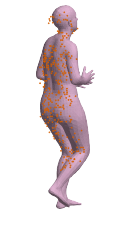}}{\includegraphics[height=1.8cm]{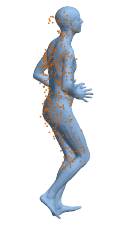}}{\includegraphics[height=1.8cm]{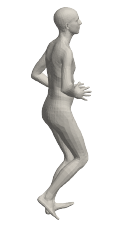}}
    & \qc{\includegraphics[height=1.8cm]{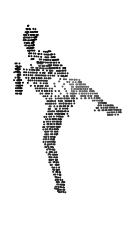}}{\includegraphics[height=1.8cm]{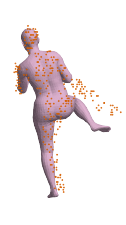}}{\includegraphics[height=1.8cm]{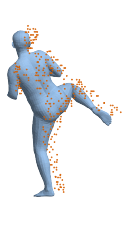}}{\includegraphics[height=1.8cm]{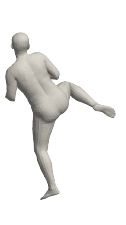}}
    & \qc{\includegraphics[height=1.8cm]{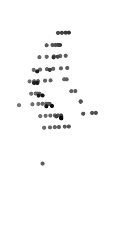}}{\includegraphics[height=1.8cm]{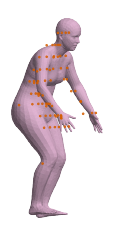}}{\includegraphics[height=1.8cm]{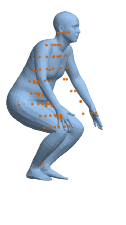}}{\includegraphics[height=1.8cm]{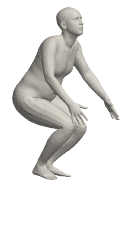}}
    & \qc{\includegraphics[height=1.8cm]{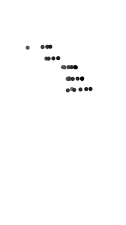}}{\includegraphics[height=1.8cm]{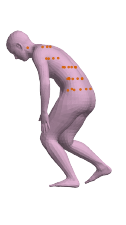}}{\includegraphics[height=1.8cm]{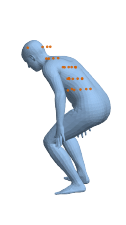}}{\includegraphics[height=1.8cm]{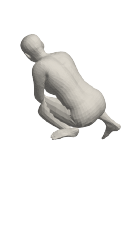}} \\[-3pt]
    {\footnotesize SLOPER4D, 1.1k pts, 50\,/\,14\,mm} & {\footnotesize FreeMotion, 967 pts, 117\,/\,31\,mm} & {\footnotesize FreeMotion, 72 pts, 89\,/\,36\,mm} & {\footnotesize\textcolor{CBorange}{failure}, 27 pts, 227\,/\,220\,mm} \\
  \end{tabular}
  \caption{\textbf{Qualitative results of a single model.}
    \emph{Top, depth cameras:} input, prediction of \family from that single frame (blue, input points in orange) and ground truth (GT, grey), with the number of points after 2\,cm voxelization and the MVE; samples with a typical error (35th--65th percentile of their set).
    \emph{Bottom, LiDAR:} comparison with LiveHPS~\cite{ren2024livehps} (released weights, 32 frames of 256 points), with the \jl of LiveHPS\,/\,ours.
    The first three frames are selected where the two models differ most (top tenth of the difference; \family is more accurate on 91\% of the SLOPER4D and 80\% of the FreeMotion frames); the last one is a failure case on which both models fail.}
  \label{fig:qual}
\end{figure*}

\paragraph{Optimization.}
We train with Muon~\cite{jordan2024muon} in the variant of Liu~\etal~\cite{liu2025muon} for the weight matrices and with AdamW~\cite{loshchilov2019adamw} for the remaining parameters.
The peak learning rate is $4.4\cdot10^{-4}$ for the encoder and decoder blocks and $4.4\cdot10^{-3}$ for all other parameters, with weight decay 0.05, a one-cycle schedule~\cite{smith2019superconvergence}, batch size 128 and mixed precision~\cite{micikevicius2018mixed}.

\paragraph{Two-stage training.}
In the first stage, the network without memory is trained on single frames for 12 epochs from random initialization.
In the second stage, the memory is added and training continues for 4 epochs at a quarter of the learning rate, on batches in which 70\% of the clouds form windows of 8 consecutive frames, 0.1--0.3\,s apart, and 30\% are independent frames; every frame of a window is supervised with its own causal history.
The two stages take 5.8 and 2.2 hours on one RTX~4070~Ti~SUPER and yield \method.
All ablations in \cref{sec:exp} are single-frame models trained as in the first stage; their final model, \base, is trained without CAPE.

\section{Experiments}
\label{sec:exp}

\begin{table}[t]
  \centering
  \small
  \setlength{\tabcolsep}{3.2pt}
  \resizebox{\linewidth}{!}{%
  \begin{tabular}{@{}lcccc@{}}
    \toprule
    Method & \jl/\vl\dn & \jg/\vg\dn & PA-MPJPE\dn & \Ang\dn \\
    \midrule
    \multicolumn{5}{@{}l}{\emph{PointHPS protocol}~\cite{cai2023pointhps}} \\
    VoteHMR~\cite{liu2021votehmr} & 81.1/97.6 & -- & 63.4 & -- \\
    PointHPS~\cite{cai2023pointhps} & 70.1/81.9 & -- & 52.0 & -- \\
    LiveHPS++~\cite{ren2024livehpspp}\textsuperscript{a} & 71.4/92.4 & 95.6/113.1 & -- & 21.47 \\
    Sparkle~\cite{ren2026sparkle} & 63.5/76.4 & 87.5/97.6 & -- & 12.60 \\
    \midrule
    \multicolumn{5}{@{}l}{\emph{MetricPose protocol}~\cite{zhu2026metricpose}\textsuperscript{b}} \\
    SAM 3D Body~\cite{yang2026sam3dbody}, RGB & 41.2/-- & 76.7/-- & 31.0 & -- \\
    MetricPose~\cite{zhu2026metricpose}, RGB-D & 37.9/-- & 39.7/-- & 30.4 & -- \\
    \midrule
    \multicolumn{5}{@{}l}{\emph{Held-out subjects (ours)}} \\
    Synthetic sources only & 58.6/72.4 & 62.0/71.5 & 44.3 & 13.41 \\
    Single-token baseline & 38.7/48.8 & 43.7/51.1 & 29.3 & 10.61 \\
    \base & 35.2/44.8 & 37.4/44.6 & 27.6 & 10.22 \\
    \method, 256 points\textsuperscript{c}, no history & 54.0/67.4 & 60.8/70.5 & 38.3 & 12.10 \\
    \method, no history & 34.7/44.3 & 36.8/\best{43.9} & 27.1 & 10.06 \\
    \method & \best{34.7}/\best{44.3} & \best{36.7}/44.0 & \best{27.1} & \best{10.00} \\
    \textcolor{gray}{Label floor} & \textcolor{gray}{3.9/10.8} & \textcolor{gray}{3.6/10.5} & \textcolor{gray}{3.5} & \textcolor{gray}{5.58} \\
    \bottomrule
  \end{tabular}}
  \caption{\textbf{Real Azure Kinect clouds of HuMMan} (SMPL; mm, degrees).
    Prior work follows PointHPS~\cite{cai2023pointhps} (a subset of clips with an unpublished split) or uses the held-out split of MetricPose; we hold out 42 subjects, so the blocks are not directly comparable.
    \emph{Label floor}: our SMPL-X labels against the original SMPL annotation.
    VoteHMR, PointHPS: MPJPE/PVE of~\cite{cai2023pointhps}, as in~\cite{ren2026sparkle}.
    \textsuperscript{a}From~\cite{ren2026sparkle}.
    \textsuperscript{b}Images with oracle person boxes; MPJPE relative to the pelvis (\jl column) and without alignment (\jg column), as evaluated in~\cite{zhu2026metricpose}, joint set not stated.
    \textsuperscript{c}Every cloud reduced to 256 points by farthest point sampling, the input of LiveHPS++ and Sparkle (both with 32 frames).}
  \label{tab:humman}
\end{table}

\begin{table}[t]
  \centering
  \small
  \setlength{\tabcolsep}{3.5pt}
  \resizebox{\linewidth}{!}{%
  \begin{tabular}{@{}llccc@{}}
    \toprule
    Method & Input & MPJPE\dn & PA-MPJPE\dn & PVE\dn \\
    \midrule
    LiDARCap~\cite{li2022lidarcap}\textsuperscript{a} & LiDAR, sequence & 144.5 & 106.2 & 177.0 \\
    LEIR~\cite{yan2024reli11d} & LiDAR, \fxp{16}{512} & 59.9 & 48.2 & -- \\
    LEIR~\cite{yan2024reli11d} & LiDAR\,$+$\,RGB\,$+$\,event & 49.2 & 40.9 & -- \\
    \midrule
    LEIR, authors' weights, our evaluation\textsuperscript{b} & LiDAR, \fxp{16}{512} & 50.5 & 40.5 & 62.6 \\
    \midrule
    \method, no history & LiDAR, \fxp{1}{512} & \best{25.9} & \best{23.1} & \best{33.9} \\
    \bottomrule
  \end{tabular}}
  \caption{\textbf{RELI11D} under its official protocol (mm): SMPL poses scored with zero shape coefficients, 24 joints relative to the root joint (MPJPE), after a similarity alignment (PA-MPJPE), and 6{,}890 vertices (PVE), on the 3{,}600 frames of the six official test files.
    First block: published results of~\cite{yan2024reli11d}.
    \textsuperscript{a}As retrained in~\cite{yan2024reli11d}.
    \textsuperscript{b}Weights distributed with the dataset, official evaluation code.}
  \label{tab:reli}
\end{table}

\begin{table}[t]
  \centering
  \small
  \setlength{\tabcolsep}{3pt}
  \resizebox{\linewidth}{!}{%
  \begin{tabular}{@{}llcc@{}}
    \toprule
    Method & Input & BEDLAM2 & FORM-HOI \\
    \midrule
    \multicolumn{4}{@{}l}{\emph{BEDLAM2 test protocol}~\cite{tesch2025bedlam2}: PA-MPJPE/MPJPE/PVE\dn} \\
    GVHMR~\cite{shen2024gvhmr} & RGB video & 36.4/62.0/52.0 & -- \\
    PromptHMR~\cite{wang2025prompthmr} & RGB video & 32.0/55.7/47.5 & -- \\
    \method, no history & depth, \fxp{1}{all} & \best{23.4}/\best{38.6}/\best{32.0} & 23.5/39.7/33.3 \\
    \midrule
    \multicolumn{4}{@{}l}{\emph{MetricPose protocol}~\cite{zhu2026metricpose}: PA-MPJPE/MPJPE/camera-frame MPJPE\dn} \\
    SAM 3D Body~\cite{yang2026sam3dbody} & RGB & 38.0/63.8/266.7 & -- \\
    MetricPose~\cite{zhu2026metricpose} & RGB-D & 30.2/50.4/166.1 & -- \\
    \method, no history & depth, \fxp{1}{all} & \best{20.1}/\best{26.2}/\best{28.5} & 19.4/27.0/30.8 \\
    \bottomrule
  \end{tabular}}
  \caption{\textbf{BEDLAM2 and FORM-HOI} in the protocols of recent methods (mm).
    Published results are on other test data: the official BEDLAM2 test set (unseen bodies and motions in new scenes; methods trained on BEDLAM and BEDLAM2~\cite{tesch2025bedlam2}) and the held-out BEDLAM2 split of MetricPose (oracle person boxes).
    Ours: the held-out subjects of our BEDLAM2 validation set and the nine held-out persons of FORM-HOI, whose ground truth is our SMPL-X fit to its MHR annotation.
    BEDLAM2 test protocol: the 127 joints and landmarks and the 10{,}475 vertices of SMPL-X after subtracting the midpoint of the hips.
    MetricPose protocol: joints relative to the pelvis, after a similarity alignment and in the camera frame; for \method the 24 SMPL joints, as in SAM 3D Body.}
  \label{tab:b2}
\end{table}

\begin{table}[t]
  \centering
  \small
  \setlength{\tabcolsep}{3.5pt}
  \resizebox{\linewidth}{!}{%
  \begin{tabular}{@{}llccc@{}}
    \toprule
    Method & Input & MPJPE\dn & PA-MPJPE\dn & Camera-frame\dn \\
    \midrule
    SAM 3D Body~\cite{yang2026sam3dbody} & RGB & 57.5 & 32.1 & 97.4 \\
    MetricPose~\cite{zhu2026metricpose} & RGB-D & 46.8 & 29.7 & 47.5 \\
    \midrule
    \method, no history & depth, \fxp{1}{all} & 33.4 & 27.0 & 43.0 \\
    \quad subjects unseen in training & depth, \fxp{1}{all} & 33.3 & 27.2 & 41.3 \\
    \bottomrule
  \end{tabular}}
  \caption{\textbf{EgoBody} in the protocol of MetricPose (mm): MPJPE relative to the pelvis, after a similarity alignment (PA-MPJPE) and in the camera frame without alignment.
    Published results: the held-out EgoBody split of MetricPose, images with oracle person boxes, joint set not stated, as evaluated in~\cite{zhu2026metricpose}.
    Ours: the official test split, clouds of the five third-person Azure Kinects (121{,}714 clouds, 66{,}648 of them of subjects that never appear in training), the 24 SMPL joints.}
  \label{tab:egobody}
\end{table}

\begin{table*}[t]
  \centering
  \small
  \setlength{\tabcolsep}{4.5pt}
  \resizebox{\textwidth}{!}{%
  \begin{tabular}{@{}clcccccccccc@{}}
    \toprule
    & & \multicolumn{2}{c}{Synthetic} & \multicolumn{2}{c}{Depth cameras} & \multicolumn{2}{c}{LiDAR} & \multicolumn{4}{c}{All nine test sets} \\
    \cmidrule(lr){3-4}\cmidrule(lr){5-6}\cmidrule(lr){7-8}\cmidrule(l){9-12}
    & Model & MVE\dn & PA-MVE\dn & MVE\dn & PA-MVE\dn & MVE\dn & PA-MVE\dn & MVE\dn & $\Delta$ & PA-MVE\dn & JAE\dn \\
    \midrule
    1 & One token, direct regression, $\ell_1$ losses, $SO(3)$ rotations & 29.5 & 17.1 & 43.3 & 26.9 & 52.4 & 27.7 & 43.3 & \dt{--} & 25.0 & 7.82 \\
    2 & $+$ 24 query tokens, shared joint head (\cref{sec:method:tokens}) & 26.5 & 16.5 & 42.8 & 26.6 & 50.8 & 27.2 & 41.8 & \dt{$-1.5$} & 24.6 & 7.70 \\
    3 & $+$ voxel voting for the pelvis (\cref{sec:method:vote}) & 26.0 & 16.4 & 41.6 & 26.8 & 49.2 & 27.4 & 40.7 & \dt{$-1.1$} & 24.7 & 7.72 \\
    4 & $+$ $\mathcal{L}_G$ and $\mathcal{L}_S$ (global rotations, voxel parts) & 23.7 & 15.5 & 39.4 & 26.0 & 46.8 & 26.8 & 38.4 & \dt{$-2.3$} & 23.9 & 7.66 \\
    5 & $+$ Frobenius form of $\mathcal{L}_R$ & 23.6 & 15.3 & 39.3 & 25.9 & 46.8 & 26.6 & 38.3 & \dt{$-0.1$} & 23.8 & 7.64 \\
    6 & $+$ Frobenius form of $\mathcal{L}_G$ & 23.6 & 15.2 & 39.1 & 25.8 & 46.7 & 26.7 & 38.2 & \dt{$-0.1$} & 23.7 & 7.63 \\
    7 & $+$ yaw--tilt instead of $SO(3)$ rotations (\base) & \best{21.2} & \best{14.4} & \best{38.0} & \best{25.4} & \best{43.3} & \best{25.9} & \best{36.0} & \dt{$-2.2$} & \best{23.1} & \best{7.47} \\
    \bottomrule
  \end{tabular}}
  \caption{\textbf{Ablation on network design: one change per row, on the full mixture.}
    Seven models, each trained for 12 epochs on the nine sources of \cref{tab:data} with the same optimizer and schedule; a row differs from the row above only in the named component ($\Delta$: change of MVE).
    SMPL-X errors in mm (JAE in degrees), averaged over the test sets of each sensor group and over all nine.}
  \label{tab:ladder}
\end{table*}

\subsection{Experimental setup}
\label{sec:exp:setup}

\paragraph{Datasets.}
We evaluate on nine test sets, one per source of the ablations, grouped by sensor: \emph{synthetic} (the official AMASS test subsets, ray-cast from fixed virtual cameras, and the BEDLAM2 validation subjects), \emph{depth cameras} (the official test splits of EgoBody and BEHAVE and the held-out subjects of HuMMan and FORM-HOI) and \emph{LiDAR} (the LiveHPS test splits of SLOPER4D and FreeMotion and the official test split of RELI11D).
All clouds with at least 10 points are evaluated; \method is additionally evaluated on 8-frame windows of the same sets.

\paragraph{Metrics.}
On our test sets, we report SMPL-X errors: the mean vertex error without alignment (MVE, mm), which includes the error of the global position, its Procrustes-aligned counterpart (PA-MVE), and the mean geodesic error of the 22 predicted rotations (JAE, degrees).
For the comparison with published methods, we follow the LiveHPS protocol~\cite{ren2024livehps,ren2024livehpspp,ren2026sparkle,an2026bmlicap} on SMPL~\cite{loper2015smpl}, with the original SMPL annotations as ground truth and the predicted SMPL-X mesh converted to the SMPL topology by the standard deformation transfer.
\jl and \vl denote the mean joint and vertex errors after subtracting the root joint, without rotational alignment; \jg and \vg retain the translation; \Ang is the angle error of the official LiveHPS code; PA-MPJPE applies a similarity alignment~\cite{umeyama1991least}.
Our SMPL-X labels deviate from the SMPL annotations by 3--4\,mm in \jl, which bounds the achievable error.

\paragraph{Comparability.}
LiveHPS, LiveHPS++ and Sparkle process 32-frame sequences in which every frame is subsampled to 256 points, whereas \method takes all points of the current frame (a median of 0.8k on SLOPER4D and 0.26k on FreeMotion) and at most seven previous frames; we also report it without history and with the 256-point input.
Prior methods are trained on the LiDAR datasets and synthetic scans, \method on the ten sources of \cref{tab:data}; \cref{sec:exp:data} isolates the contribution of the data.
To validate our test files and metric implementation, we re-evaluate the released LiveHPS model with its official script (\cref{tab:lidar}): it reproduces the published \jl on SLOPER4D (47.4 \vs 48.3\,mm) but obtains 52.8 instead of the published 69.4\,mm on the released FreeMotion test file.
Published FreeMotion results may thus not be directly comparable with results on this file, and we regard that comparison as indicative.

\subsection{Comparison with state-of-the-art methods}
\label{sec:exp:sota}

\paragraph{LiDAR benchmarks.}
As shown in \cref{tab:lidar}, \method achieves a \jl of 28.6\,mm on SLOPER4D and 41.8\,mm on FreeMotion with a causal history of seven frames.
It outperforms the best published results, those of the 32-frame BMLiCap, by 7.9 and 5.4\,mm, and Sparkle by 12.8 and 17.2\,mm.
On the same test files, the released LiveHPS model is 18.8 and 11.0\,mm behind (\cref{fig:qual}), and still 10.0\,mm on SLOPER4D when it is given its preferred input frame.
The angle error of \method is more than a quarter below that of Sparkle, and its global error is 38\,mm lower on FreeMotion and on par on SLOPER4D, where 26 near-empty frames account for over half of it.

We further analyze where this margin originates.
\emph{(i) Input.}
Restricted to the 256 points per frame of prior pipelines, without retraining, \method obtains 36.8 and 48.2\,mm, still 4.6 and 10.8\,mm below Sparkle and within 0.3 and 1.0\,mm of the 32-frame BMLiCap; processing all points thus accounts for about 8 and 6\,mm, at no additional cost for a sparse-voxel encoder (\cref{sec:exp:sparsity}).
\emph{(ii) Data.}
A plain single-token regressor trained on our mixture already achieves 31.3 and 49.8\,mm, whereas \base trained without LiDAR sources yields 48.9 and 82.6\,mm.
\emph{(iii) Design.}
The proposed components reduce the error of the single-token baseline to 29.0 and 42.4\,mm (\base).
\emph{(iv) History.}
Without its history, \method obtains 28.9 and 42.9\,mm; temporal context thus contributes only 0.3 and 1.1\,mm (\cref{sec:exp:memory}).

\paragraph{RELI11D.}
\cref{tab:reli} follows the evaluation code released with RELI11D~\cite{yan2024reli11d}, which scores SMPL poses with zero shape coefficients on test windows of 16 frames with 512 points per frame.
We convert the SMPL-X prediction of \method into an SMPL pose by fitting SMPL to the predicted mesh mapped to the SMPL topology; applied to our labels, this conversion alone yields an MPJPE of 4.5\,mm.
With the weights distributed with the dataset, LEIR obtains an MPJPE of 50.5\,mm on the same test files, below the published 59.9\,mm.
From single frames of the same 512 points, \method obtains 25.9\,mm, about half the error of LEIR, including its variant with RGB and event cameras.

\paragraph{Depth cameras.}
On the Azure Kinect clouds of the 42 held-out HuMMan subjects (\cref{tab:humman}), \method achieves a \jl of 34.7\,mm and a PA-MPJPE of 27.1\,mm, with or without its history; the global error exceeds the local one by only 2\,mm, \ie the body is accurately localized on dense clouds.
Point-cloud methods report at best 63.5 and 52.0\,mm in the protocol of PointHPS, whose split of clips is not public, so these results are not directly comparable; we note that our evaluation holds out all test subjects and that our model trained on synthetic sources alone already achieves 58.6\,mm.
Reduced to the 256 points per frame of Sparkle, \method obtains 54.0\,mm from single frames, below the 63.5\,mm that Sparkle obtains from 32 frames.
The lowest published errors on HuMMan, those of the RGB-D method MetricPose~\cite{zhu2026metricpose} on its own held-out split (37.9, 30.4 and 39.7\,mm relative to the pelvis, after a similarity alignment and in the camera frame), are also above those of \method (34.7, 27.1 and 36.7\,mm).

\paragraph{EgoBody.}
On the official test split of EgoBody, \cref{tab:egobody} uses the metrics of MetricPose.
From single clouds of the five third-person Azure Kinects, \method obtains 33.4, 27.0 and 43.0\,mm relative to the pelvis, after a similarity alignment and in the camera frame, and 33.3, 27.2 and 41.3\,mm on the subjects that never appear in training, against 46.8, 29.7 and 47.5\,mm reported for MetricPose with RGB-D input on its own held-out split.

\paragraph{BEDLAM2 and FORM-HOI.}
The official BEDLAM2 test set is released without depth maps, and FORM-HOI has no published results.
\cref{tab:b2} therefore scores our held-out subjects of both sets in the protocols of the BEDLAM2 test set~\cite{tesch2025bedlam2} and of MetricPose; test data and input differ from those of the published results, so the comparison is indicative.
From single depth frames, \method obtains an MPJPE of 38.6\,mm in the protocol of the BEDLAM2 test set, against 55.7\,mm for PromptHMR on RGB video, and a camera-frame error of 28.5\,mm, against 166.1\,mm for MetricPose with RGB-D input.

\subsection{Ablation on network design}
\label{sec:exp:ablation}

\begin{table}[t]
  \centering
  \small
  \setlength{\tabcolsep}{3.4pt}
  \resizebox{\linewidth}{!}{%
  \begin{tabular}{@{}lcccccc@{}}
    \toprule
    & \multicolumn{2}{c}{Synthetic} & \multicolumn{2}{c}{Depth cameras} & \multicolumn{2}{c}{LiDAR} \\
    \cmidrule(lr){2-3}\cmidrule(lr){4-5}\cmidrule(l){6-7}
    Training sources & MVE\dn & PA\dn & MVE\dn & PA\dn & MVE\dn & PA\dn \\
    \midrule
    Synthetic (2) & \best{20.5} & \best{14.1} & 60.6 & 39.4 & 88.9 & 52.4 \\
    $+$ depth cameras (6) & 21.2 & 14.3 & \best{37.5} & 25.3 & 84.8 & 49.3 \\
    $+$ LiDAR (9, \base) & 21.2 & 14.4 & 38.0 & 25.4 & \best{43.3} & \best{25.9} \\
    $+$ CAPE (10) & 20.8 & 14.4 & 37.9 & \best{25.2} & 43.5 & 26.0 \\
    \bottomrule
  \end{tabular}}
  \caption{\textbf{Ablation on training data.}
    The single-frame model trained on growing subsets of the sources; in the first three rows the epoch size is fixed and the draws of the removed sources go to BEDLAM2 and AMASS.
    Mean MVE and PA-MVE (mm) per sensor group.}
  \label{tab:data_ladder}
\end{table}

We start from a plain regressor and add one component at a time, training every variant on the nine-source mixture (no CAPE, no memory) for 12 epochs (\cref{tab:ladder}).
The baseline (row 1) applies the same encoder and decoder blocks to a single token, with linear heads, $\ell_1$ losses and uniform $SO(3)$ rotation augmentation.

\emph{Query tokens (row 2).}
Separate tokens for the shape, the translation and every joint, decoded by a shared joint head, reduce MVE by 1.5\,mm and PA-MVE by 0.4\,mm at a nearly unchanged number of parameters, with the largest gain on the synthetic sets ($-3.0$\,mm).
\emph{Voxel voting (row 3)} reduces MVE by a further 1.1\,mm ($-1.6$\,mm on the LiDAR sets) while PA-MVE remains unchanged, which confirms that it improves the localization of the body rather than its pose.
\emph{Auxiliary supervision (row 4).}
The global-rotation loss and the voxel part segmentation yield the largest single improvement, $-2.3$\,mm MVE and $-0.8$\,mm PA-MVE, consistently across sensor groups; as both were added in one run, their individual contributions cannot be separated.
\emph{Loss form (rows 5--6).}
Replacing the element-wise $\ell_1$ rotation losses with the Frobenius form changes the means by only 0.1\,mm per step, which is within run-to-run variation; we claim no gain and adopt this form because it depends on the error angle alone.
\emph{Gravity-aware augmentation (row 7).}
Replacing uniform $SO(3)$ rotations with the yaw--tilt augmentation reduces MVE by 2.2\,mm and benefits every group, most notably the sparse FreeMotion clouds ($-8.0$\,mm) and the synthetic sets ($-2.4$\,mm).

Overall, the chain reduces MVE by 7.3\,mm (17\%) and PA-MVE by 1.9\,mm without adding a block.

\subsection{Ablation on training data}
\label{sec:exp:data}

\cref{tab:data_ladder} reports the single-frame model trained on growing subsets of the sources.
\emph{Each sensor family requires its own data.}
Adding real depth-camera data reduces the error on the depth-camera test sets from 60.6 to 37.5\,mm but the LiDAR error only from 88.9 to 84.8\,mm; adding LiDAR data then halves the latter to 43.3\,mm.
The gap is not explained by density alone, since the model without LiDAR sources is already trained with depth clouds subsampled to LiDAR density (\cref{sec:training}).
\emph{Joint training costs little.}
Adding the four real depth sources increases the synthetic error by 0.7\,mm, and adding the three LiDAR sources increases the depth-camera error by 0.5\,mm and leaves the synthetic error unchanged.
CAPE as a tenth source changes no group mean by more than 0.4\,mm and yields 24.7\,mm MVE on its own held-out subjects.
A single set of weights thus stays within 0.7\,mm of the models specialized to the synthetic and depth-camera test sets.

\subsection{Effect of the token memory}
\label{sec:exp:memory}

\begin{table}[t]
  \centering
  \small
  \setlength{\tabcolsep}{4pt}
  \begin{tabular}{@{}llcccc@{}}
    \toprule
    & & \multicolumn{3}{c}{8-frame windows} & Thinned \\
    \cmidrule(lr){3-5}\cmidrule(l){6-6}
    Model & History & MVE\dn & JAE\dn & Accel\dn & MVE\dn \\
    \midrule
    After stage 1 & -- & 35.0 & 8.26 & 1.065 & -- \\
    \method & off & 34.9 & 8.22 & 1.069 & 77.7 \\
    \method & 7 frames & \best{34.5} & \best{8.09} & \best{1.008} & \best{69.4} \\
    \bottomrule
  \end{tabular}
  \caption{\textbf{Effect of the token memory.}
    Means over windows of nine test sets (frames 0.2\,s apart; mm, degrees, m/s$^2$): the network after the first training stage, and \method without and with its history.
    \emph{Thinned}: the current frame keeps 10\% of its points, the history is intact.}
  \label{tab:memory}
\end{table}

\cref{tab:memory} compares \method with the single-frame network obtained after the first training stage.
With its history disabled, \method matches this network on the windows and on the mean of the nine single-frame test sets (36.0\,mm); the second stage thus preserves the single-frame accuracy on average, although the error on the AMASS test set increases from 11.7 to 13.0\,mm.
With the history, MVE, JAE and the acceleration error~\cite{kanazawa2019hmmr} decrease by 1.4\%, 2.1\% and 5.3\% with respect to the first-stage network, on 8, 9 and 9 of the nine sets.
The gain depends on the quality of the current frame: 0.7\% for clouds of more than 1000 points, 3--5\% for clouds of fewer than 200 points, and 11\% when the current frame is thinned to a tenth of its points.
On top of an encoder that processes all points, temporal context therefore mainly improves smoothness and robustness to degraded frames rather than accuracy.

\subsection{Robustness and efficiency}
\label{sec:exp:sparsity}

\paragraph{Sparse clouds.}
On FreeMotion, \method stays below 36\,mm \jl for clouds of at least 200 points and reaches 63\,mm at 50--100 points.
Keeping sparse scans in training is essential: a single-frame model trained without them matches \base above 200 points but reaches only 96\,mm at 50--100 points and flips the body (\jl above 300\,mm) in 3.1\% of them, against 0.7\%.
Below 50 points (2.4\% of FreeMotion), \method obtains 173\,mm and flips 7\% of the clouds (13\% without its history).

\paragraph{Efficiency.}
\method has 50.0M parameters, about half the size of the released LiveHPS model (105.7M) and below the 82.3M reported for Sparkle~\cite{ren2026sparkle}; PointNet++-based single-frame models are smaller (PointHPS: 4.7M~\cite{cai2023pointhps}).
On an RTX~4090, a single cloud takes 10\,ms including the SMPL-X layer, for dense and sparse clouds alike, and the memory adds 6--29\% for an 8-frame window.

\section{Conclusion}
\label{sec:conclusion}
We have presented \family, a unified network for human pose and shape estimation from range data.
A LitePT encoder with per-joint query tokens, a shared joint head, voxel voting and a causal token memory, trained on ten sources in one label space, serves depth cameras and LiDAR with a single set of weights.
A single frame already surpasses published sequence-based methods on LiDAR benchmarks, while the memory adds smoothness and robustness.

\paragraph{Limitations.}
\label{sec:limits}
\family estimates one person per cloud and requires this person to be segmented from the scene, as do all compared methods; on several datasets the crop relies on released masks or labels.
It requires the vertical axis to within about $15^\circ$, does not estimate hands or the face, and often fails below 50 points.
The comparison with published methods is not fully controlled: we train on more sources, the 256-point result uses weights not trained for this input, the released FreeMotion test file does not reproduce the published LiveHPS result, and the HuMMan splits differ.

{
  \small
  \bibliographystyle{ieeenat_fullname}
  \bibliography{main}
}
\end{document}